\documentclass[nohyperref]{article}

\usepackage{microtype}
\usepackage{graphicx}
\usepackage{subfigure}
\usepackage{booktabs} % for professional tables

\usepackage[hidelinks]{hyperref}

\usepackage[review]{icml2023}

\usepackage{amsmath}
\usepackage{amssymb}
\usepackage{mathtools}
\usepackage{amsthm}

\usepackage[capitalize,noabbrev]{cleveref}

\theoremstyle{plain}
\newtheorem{theorem}{Theorem}[section]

\theoremstyle{definition}
\newtheorem{definition}[theorem]{Definition}

\theoremstyle{remark}

\usepackage{amsmath}
\usepackage[textsize=tiny]{todonotes}
\newcommand{\Dd}[1]{\mathcal{D}^{#1}}

\newcommand{\Ss}{\mathcal{S}}

\DeclareMathOperator*{\argmin}{arg\,min}
\newtheorem{prop}{Proposition}

\def\1{\bm{1}}

\DeclareMathAlphabet{\mathsfit}{\encodingdefault}{\sfdefault}{m}{sl}
\SetMathAlphabet{\mathsfit}{bold}{\encodingdefault}{\sfdefault}{bx}{n}

\usepackage{url}            % simple URL typesetting

\usepackage{graphicx}

\usepackage{float}
\usepackage{amsmath}
\usepackage{multirow}
\usepackage{csquotes}
\usepackage{todonotes}
\usepackage{rotating}
\usepackage{svg}
\usepackage{caption}
\usepackage{verbatim}
\usepackage{amssymb}
\usepackage{pifont}% http://ctan.org/pkg/pifont
\newcommand{\cmark}{\ding{51}}%
\newcommand{\xmark}{\ding{55}}%
\icmltitlerunning{Explanation Shift}

\usepackage{xcolor}

\begin{document}

\twocolumn[
%\icmltitle{Explanation Shift: detecting changes on the model behaviour via out-of-distribution explanations}
\icmltitle{Explanation Shift:\\ How Did the Distribution Shift Impact the Model?}

% It is OKAY to include author information, even for blind
% submissions: the style file will automatically remove it for you
% unless you've provided the [accepted] option to the icml2022
% package.

% List of affiliations: The first argument should be a (short)
% identifier you will use later to specify author affiliations
% Academic affiliations should list Department, University, City, Region, Country
% Industry affiliations should list Company, City, Region, Country

% You can specify symbols, otherwise they are numbered in order.
% Ideally, you should not use this facility. Affiliations will be numbered
% in order of appearance and this is the preferred way.
\icmlsetsymbol{equal}{*}

\begin{icmlauthorlist}
\icmlauthor{Carlos Mougan}{soton}
\icmlauthor{Klaus Broelemann}{schufa}
\icmlauthor{David Masip}{glovo}
\icmlauthor{Gjergji Kasneci}{tubinguen}
\icmlauthor{Thanassis Thiropanis}{soton}
\icmlauthor{Steffen Staab}{soton,stuttgart}
\end{icmlauthorlist}

\icmlaffiliation{soton}{Department of Computer Science, University of Southampton,United Kingdom}
\icmlaffiliation{schufa}{Schufa Holding AG, Germany}
\icmlaffiliation{glovo}{Glovo, Barcelona, Spain}
\icmlaffiliation{tubinguen}{University of Tübingen, Germany}
\icmlaffiliation{stuttgart}{University of Stuttgart, Institute
of Parallel and Distributed Systems, Stuttgart, Germany}
\icmlcorrespondingauthor{Carlos Mougan}{c.mougan@soton.ac.uk}

% You may provide any keywords that you
% find helpful for describing your paper; these are used to populate
% the "keywords" metadata in the PDF but will not be shown in the document
\icmlkeywords{Machine Learning, ICML}

\vskip 0.3in
]

% this must go after the closing bracket ] following \twocolumn[ ...

% This command actually creates the footnote in the first column
% listing the affiliations and the copyright notice.
% The command takes one argument, which is text to display at the start of the footnote.
% The \icmlEqualContribution command is standard text for equal contribution.
% Remove it (just {}) if you do not need this facility.

%\printAffiliationsAndNotice{}  % leave blank if no need to mention equal contribution
%\printAffiliationsAndNotice{\icmlEqualContribution} % otherwise use the standard text.

\theoremstyle{definition}
\newtheorem{example}{Example}[section]

\begin{abstract}
As input data distributions evolve, the predictive performance of machine learning models tends to deteriorate.
In practice, new input data tend to come without target labels. Then, state-of-the-art techniques model input data distributions or model prediction distributions and try to understand issues regarding the interactions between learned models and shifting distributions.
We suggest a novel approach that  models how explanation characteristics shift when affected by distribution shifts.
We find that the modeling of explanation shifts can be a better indicator for detecting out-of-distribution model behaviour than state-of-the-art techniques.
We  analyze different types of distribution shifts using synthetic examples and real-world data sets. We provide an algorithmic method that allows us to inspect the interaction between data set features and learned models and compare them to the state-of-the-art. We release our methods in an open-source Python package, as well as the code used to reproduce our experiments.
\end{abstract}

\section{Introduction}
Machine learning theory gives us the means to forecast the quality of ML models on unseen data, provided that this data is sampled from the same distribution as the data used to train and evaluate the model. If unseen data is sampled from a different distribution, model quality may deteriorate.

Model monitoring tries to signal and possibly even quantify such decay of trained models. Such monitoring is challenging because only in a few applications do unseen data come with labels that allow for direct monitoring of model quality. Much more often,  deployed ML models encounter unseen data for which target labels are lacking or biased~\cite{DBLP:conf/nips/RabanserGL19,DesignMLSystems,DBLP:conf/icml/ZhangGR21}.   

%the deployment data has accessible labels, and predictive quality measures can be calculated. There are many use cases where there are no labels for the data used in deployment, or the labels are difficult to obtain.
%\steffen{Every time you say `performance' I think of run time, never of quality. Maybe this is just me. At least `performance' is ambiguous.}

%\todo{@Gjergji: Add some real-world examples here: Real-time advertisement ranking on the web or mobile apps (labels exist only partially and may be biased, i.e., the observed clicks), credit scoring, fraud detection (no labels after deployment; labels are only collected in batches and serve for the development of the next model). Also, state that most of these use cases are based on heterogeneous tabular data.}

%Hence, typically, the only data available in ML applications are labeled source data and unlabeled deployment data~\cite{garg2022leveraging}.
Detecting changes in the quality of deployed ML models in the absence of labeled data remains a challenge~\cite{DBLP:conf/aaai/RamdasRPSW15,DBLP:conf/nips/RabanserGL19}. State-of-the-art techniques model statistical distances between training and unseen data distributions~\cite{continual_learning,clouderaff} or statistical distances between distributions of  model predictions~\cite{garg2022leveraging,garg2021ratt}. The shortcomings of these measures of \emph{distribution shifts} is that they do not relate changes of distributions to how they interact with trained models.  Often, there is the need to go beyond detecting changes and understand how feature attribution changes~\cite{10.1145/3534678.3542617,mougan2022monitoring,continual_learning}.

% xAI
The field of explainable AI has emerged as a way to understand model decisions ~\cite{xai_concepts,molnar2019} and interpret the inner workings of black box models~\cite{guidotti_survey}. The core idea of this paper is to go beyond the modeling of distribution shifts and monitor for \emph{explanation shifts} to signal change of interactions between learned models and dataset features in tabular data. 
We newly define explanation shift to be constituted by the statistical comparison between how predictions from training data are explained and how predictions on new data are explained.  %Explanation shift goes beyond the mere recognition of changes in data distributions towards the recognition of changes of how data distributions relate to the models' inner workings.
 
In summary, our contributions are:
\begin{itemize}
    \item We propose measures of explanation shifts as a key indicator for investigating the interaction between distribution shift and learned models.

    \item We define an \textit{Explanation Shift Detector} that operates on the explanation space allowing for more sensitive and explainable changes of interactions between distribution shifts and learned models.
    
    \item We compare our monitoring method that is based on explanation shifts with methods that are based on other kinds of distribution shifts. We find that monitoring for explanation shifts results in better indicators for varying model behaviour.
    
 %   provide a mathematical comparison between changes on the explanation versus input and versus predictions that shows how simple, but key types of distribution shift interact with linear models such that measures of explanation shift become much better indicators of model behaviour changes than measures of distribution shift or prediction shift. 

    \item We release an open-source Python package\footnote{to be released upon acceptance}, which implements our \enquote{\textit{Explanation Shift Detector}} that is \texttt{scikit-learn} compatible~\cite{pedregosa2011scikit}, along the code and usage tutorials for further reproducibility.
    
    %\item \textcolor{red}{We compare how measures of shifts of distributions, predictions, and explanations affect the task of quantifying model degradation in terms of predictive performance. }
    
    %\item \textcolor{red}{We provide comprehensive experimental evaluations and comparisons on real-world data concerned with  detecting and quantifying model decay.}
\end{itemize}

\section{Foundations and Related Work}\label{sec:foundations}
%\gjergji{This comes quite unexpectedly. I think we should start with explanations shift detection methods and their shortcomings; we should then hint at the advantages of analyzing explanation shifts and then motivate why we choose Shapley values as a representative explanation method for our approach.}\carlos{I agree, but for the next resubmission}
\subsection{Explainable AI}
Explainability has become an important concept in legal and ethical data processing guidelines and machine learning applications ~\cite{selbst2018intuitive}. A wide variety of methods have been developed to account for the decision of algorithmic systems ~\cite{guidotti_survey,DBLP:conf/fat/MittelstadtRW19,DBLP:journals/inffus/ArrietaRSBTBGGM20}. One of the most popular approaches to explaining machine learning models has been the use of Shapley values to attribute relevance to features used by the model~\cite{lundberg2020local2global,lundberg2017unified}. The Shapley value is a concept from coalition game theory that aims to allocate the surplus generated by the grand coalition in a game to each of its players~\cite{shapley}. The Shapley value $\mathcal{S}_j$ for the $j$'th player can be defined via a value function $\mathrm{val}:2^N \to \mathbb{R}$ of players in $T$:

\begin{small}
\begin{gather}
\mathcal{S}_j(\mathrm{val}) = \sum_{T\subseteq N\setminus \{j\}} \frac{|T|!(p-|T|-1)!}{p!}(\mathrm{val}(T\cup \{j\}) - \mathrm{val}(T))
\end{gather}
\end{small}

In machine learning, $N=\{1,\ldots,p\}$ is the set of features  occurring in the training data, and $T$ is used to denote a subset of $N$. Given that
 $x$ is the feature vector of the instance to be explained, and the term $\mathrm{val}_x(T)$ represents the prediction for the feature values in $T$ that are marginalized over features that are not included in $T$:
% \begin{gather}
% \mathrm{val}_{\hat{f},x}(T) = E_{X_{N\setminus T}}[\hat{f}(X)|X_T=x_T]-E_X[\hat{f}(X)]
% \end{gather}
\begin{gather}
\mathrm{val}_{f,x}(T) = E_{X|X_T=x_T}[f(X)]-E_X[f(X)]
\end{gather}

The Shapley value framework satisfies several theoretical properties~\cite{molnar2019,shapley,WINTER20022025,aumann1974cooperative}, and our approach is based on the efficiency property. Our approach works with explanation techniques that fulfill efficiency  and uninformative properties, and we use Shapley values as an example.

\textbf{Efficiency.} Feature contributions add up to the difference of prediction for $x^{\star}$ and the expected value of $f$:
\begin{gather}
    \sum_{j \in N} \Ss_j(f, x^{\star}) = f(x^{\star}) - E[f(X)])
\end{gather}

%\klaus{I would indicate, that $\mathrm{val}$ also depends on $\hat{f}$: $\mathrm{val}_{\hat{f},x}(T)$ }
%\steffen{The first expected value does not make sense as the index in the E does have a different dimensionality than the X. And this seems to be only one bug}\klaus[inline]{I still think this equation is valid: we condition on the features in $T$, so the expectation has to be done overall feature, not in $T$. So $E_{X_{N\setminus T}}$ is the expectation we want to build. If we ignore the ordering, we could write $\hat{f}(X_T,X_{N\setminus T})$ instead of $\hat{f}(X)$, which might make it clearer, but formally less correct.}

It is important to differentiate between the theoretical Shapley values and the different implementations that approximate them. We use  TreeSHAP as an efficient implementation of an approach for tree-based models of Shapley values~\cite{lundberg2020local2global,molnar2019,Zern2023Interventional}, particularly we use the observational (or path-dependent) estimation  ~\cite{DBLP:journals/corr/abs-2207-07605,DBLP:conf/nips/FryeRF20,DBLP:journals/corr/ShapTrueModelTrueData} and for linear models we use the correlation dependent implementation that takes into account feature dependencies \cite{DBLP:journals/ai/AasJL21}.

\subsection{Related Work on Tabular Data}
Evaluating how two distributions differ has been a widely studied topic in the statistics and statistical learning literature~\cite{statisticallearning,datasetShift}, that have advanced recently in last years ~\cite{DBLP:conf/nips/ParkAKP21,DBLP:conf/nips/LeeLLS18,DBLP:conf/icml/ZhangSMW13}.~\cite{DBLP:conf/nips/RabanserGL19} provide a comprehensive empirical investigation, examining how dimensionality reduction and two-sample testing might be combined to produce a practical pipeline for detecting distribution shifts in real-life machine learning systems. 

Some techniques detect that new data is out-of-distribution data when using neural networks based on the prediction space~\cite{fort2021exploring,NEURIPS2020_219e0524}. They use the maximum softmax probabilities/likelihood as a confidence score~\cite{DBLP:conf/iclr/HendrycksG17}, temperature or energy based scores ~\cite{DBLP:conf/nips/RenLFSPDDL19,DBLP:conf/nips/LiuWOL20,DBLP:conf/nips/WangLBL21}, they extract information from the gradient space~\cite{DBLP:journals/corr/GradientShift}, they fit a Gaussian distribution to the embedding or they use the Mahalanobis distance for out-of-distribution detection~\cite{DBLP:conf/nips/LeeLLS18,DBLP:journals/corr/reliableShift}. 

Many of these methods are  developed specifically for neural networks that operate on image and text data, and often they can not be directly applied to traditional machine learning techniques. 
For image and text data, one may build on the assumption that the relationships between relevant predictor variables ($X$) and
response variables ($Y$) remains unchanged, i.e.\ that no \emph{concept shift} occurs. For instance, the essence of how a dog looks like remains unchanged over different data sets, even if contexts may change. Thus, one can define  invariances on the latent spaces of deep neural models, which are not applicable to tabular data in a likewise manner. For example, predicting buying behavior before, during, and after the COVID-19 pandemic constitutes a concept shift that is not amenable to such methods.
 We  focus on such tabular data where techniques such as gradient boosting decision trees achieve state-of-the-art model performance~\cite{grinsztajn2022why,DBLP:journals/corr/abs-2101-02118,BorisovNNtabular}. 
 %The explanation space is a projection of a machine-learning model to a distribution space that has more dimensions than the output space and has theoretical properties that we exploit in contrast to latent spaces.

%\steffen{The following is rather unclear and lacks citation.}Other research lines, such as membership inference (or membership classifiers) aims to detect that given a data record and a machine learning model, detect if the record was in the model’s training dataset or not ~\cite{DBLP:conf/sp/ShokriSSS17}. Our approach is slightly different, as we don't attempt to predict if a certain instance belongs or not to the training data, but if its distribution is as the training dataset.  \steffen{I do not understand why the topic of discovering that an instance was part of the training data is even an issue that is worth discussing.}

% For tabular data libraries such as \cite{alibi-detect} recolect the work 
% Shap original paper suggestion

The first approach of using explainability to detect changes in the model was suggested by~\cite{lundberg2020local2global} who monitored the SHAP value contribution in order to identify possible bugs in the pipeline. A similar approach by~\cite{li2022enabling} allows for tracking distributional shifts and their impact among input variables using slidSHAP a novel method for unlabelled data streams. 
%Other works use explainable AI towards distribution shifts are 
\cite{DBLP:conf/aistats/BudhathokiJBN21}  identifies the drivers of distribution changes using graphical causal models and feature attributions using Shapley values. % The key idea is that, given a causal graph, they factorize the joint distribution into independent causal conditionals. Then, changes in the joint distribution can  be attributed to changes in some of the causal conditionals.
  In our approach we do not rely on additional information, such as a causal graph~\cite{schrouff2022diagnosing}. We will analyse later why monitoring changes in the input data distributions and the prediction distributions are not sufficient to monitor for change.
%\steffen{One question that was not answered in the past (and I have not read the complete paper yet) was: what are the best competing methods that will be compared to empirically. This seems like a worthy candidate to compare to.}\carlos{It's on the synthetic data experiments part.}\steffen{How was your AAAI2023 paper related?}

%Other lines have aimed to predict or find proxies for OOD model performance using agreement-on-the-line \cite{baek2022agreementontheline}, accuracy on the line~\cite{DBLP:conf/icml/MillerTRSKSLCS21}, explainable uncertainty\cite{mougan2022monitoring} or, distribution-free uncertainty quantification under covariate shift \cite{prinster2022jaws}\steffen{how are they related to our approach? Terms like `agreement-on-the-line' do not explain anything. Our paper must be self-contained and not require to read other papers.}

%\carlos{not sure how to fit with the next paragraph.} % Then we provide an experiment where we can see that changes on the explanations serve as a better model degradation proxy than experiments on input data or model predictions.

Recent work confirms these principal limitations by theorems about the impossibility to predict model degradation~\cite{garg2022leveraging,chen2022estimating} or the impossibility to detect that new data is out-of-distribution~\cite{fang2022is,DBLP:conf/icml/ZhangGR21,guerin2022outofdistribution}.
We do not overcome such limitations, however, our approach provides for hints that allows the machine learning engineer to better understand the change of interactions resulting from shifting data distributions and learned models.

%focused on delimiting the results that can be obtained in the field, from stating impossibility theorems delimiting the situations in which we can predict model performance~\cite{garg2022leveraging,chen2022estimating} or out-of-distribution data detection~\cite{fang2022is,DBLP:conf/icml/ZhangGR21,guerin2022outofdistribution}. Our work is limited to recognizing that changes in the explanations are key indicators for changes in the model behaviour in situations where we have unlabeled OOD data.

%\subsection{Related Work on Images and Text}
%
%We study the problems possibly incurred by distribution shifts on tabular data, which still constitutes a major field of application for machine learning. In contrast, most recent research on model monitoring proposes methods for modalities such as texts or images. These recent methods build on the assumption that the relationships between relevant predictor variables ($X$) and response variables ($Y$) remains unchanged, i.e.\ that no \emph{concept shift} occurs. For instance, their assumption is that the essence of how a dog looks like remains unchanged over different contexts. Thus, they can define  invariances on the latent spaces of deep neural models, which are not applicable to tabular data in a likewise manner. For example, predicting buying behavior before, during, and after the COVID-19 pandemic constitutes a concept shift that is not amenable to such methods.
\section{Methodology}
\subsection{Key Terminology}

The objective of supervised learning is to induce a function $f_\theta:\mbox{dom}(X) \to \mbox{dom}(Y)$, where $f_\theta$ is from a family of functions $f_\theta \in F$, from training set  $\Dd{tr}=\{(x_0^{tr},y_0^{tr})\ldots, (x_n^{tr},y_n^{tr})\} \subseteq \mbox{dom}({X} \times {Y})$
with  predictor variables $X$ and target variable $Y$, respectively. The estimated hypothesis $f_\theta$ is expected to generalize well on  new, previously unseen data $\Dd{new}=\{x_0^{new}, \ldots, x_k^{new} \} \subseteq \mbox{dom}(X)$
, for which the target labels are unknown. The traditional machine learning assumption is that training data $\Dd{tr}$ and novel data  $\Dd{new}$ are sampled from the same underlying distribution $\mathbf{P}(X\times Y)$.

%\gjergji{You first use $X$ and $Y$ as the domain and range of $f$. In the previous sentence you use them as variables in $P(X,Y)$. You can use $\mathcal{X}, \mathcal{Y}$ as the domain and range of $f$ instead. For a distribution, I would use lower-case $p$, otherwise (with $P$ as a probability measure conventionally) it is just a probability.}

%Then we call \textit{model behaviour} to the procedure of how the input data contributes to the prediction with respect to the model $f_\theta$. Then, out-of-distribution model behaviour detection is a binary classification problem that relies on a score to differentiate between in- and out-of-distribution samples that affect the model predictions.

%If we have a hold-out test data set $\Dd{te}=\{(x_0^{te},y_0^{te})\ldots, (x_m^{te},y_m^{te})\} \subseteq \mathcal{X} \times \mathcal{Y}$ disjoint from $\Dd{tr}$, but also sampled from $P(X,Y)$, one may use $\Dd{te}$ to estimate performance indicators for $\Dd{new}$. Commonly, novel data is sampled from a  distribution $P'(X,Y)$ that is different from $P(X,Y)$.% and there is no hold-out data set sampled from $P'(X,Y)$ to estimate model performance. We use $\Dd{ood} \sim P'(X,Y)$ to refer to such novel, out-of-distribution data.

\begin{definition}[Out-of-distribution data]
Given a  training data set  $\Dd{tr}=\{(x_0^{tr},y_0^{tr})\ldots, (x_n^{tr},y_n^{tr})\}\sim \mathbf{P}(X\times Y)$
and $\Dd{new}=\{x_0^{new},\ldots, x_k^{new}\}\sim \mathbf{P}(X')$, we say that $\Dd{new}$ is out-of-distribution if $\mathbf{P}(X)$ and $\mathbf{P}(X')$ are different distributions.
%\gjergji{Careful: A probability cannot be generated. There is a probability of an event under a given distribution over events. If $P$ is supposed to represent a distribution with a prior distribution, you should use $p$ instead. But in general, *data* is generated by some distribution.}
\end{definition}

\begin{definition}[Out-of-distribution predictions]
Given a model $f_\theta:\mbox{dom}(X) \to \mbox{dom}(Y)$ with parameters $\theta$ learned from training set  $\Dd{tr}=\{(x_0^{tr},y_0^{tr})\ldots, (x_n^{tr},y_n^{tr})\}$, we say that
$\Dd{new}=\{x_0^{new},\ldots, x_k^{new}\}$
 has out-of-distribution predictions with respect to model $f_\theta$ if $f_\theta(\Dd{tr}_{X})$ is sampled from a  distribution different than $f_\theta(\Dd{new})$.
\end{definition}

\begin{definition}[Explanation Space]
An explanation function $\mathcal{S}:F\times \mbox{dom}(X)\to \mathbb{R}^p$ maps a model $f_\theta$  and data of interest $x\in \mathbb{R}^p$ to a vector of contributions $\mathcal{S}(f_\theta, x)\in \mathbb{R}^p$. Given a dataset $\mathcal{D}$, its explanation space is the matrix with rows $\Ss(f_\theta, x_i)^\top$ for $x_i \in \mathcal{D}$. 

%\klaus{I would distinguish between \emph{explanation function} and \emph{SHAP Values}, which is a specific explanation function (Sabaas would be another). Note, that these are not Shapley Values, but SHAP Values with this signature.}
\end{definition}\label{def:explanationSpace}

We use  Shapley values to define the explanation function $\Ss$. 
%\steffen{What is the signature?}\carlos{can i say $dim(x_i) \equiv dim(\mathcal{S}(f_\theta,x_i))$} \klaus{You could write something like $\mathcal{S}:F\times \mathcal{X}\to \mathbb{R}^p$} \klaus[inline]{Might be better to define it for a single feature vector $x$. Batch computation is an implementation detail.}\klaus{Be nice to the reader. $S$ (the subsets) and $\mathcal{S}$ (the explanation function) look quite similar. Consider to use e.g. $T$ for the former one.}
\begin{definition}[Out-of-distribution explanations]
Given a model $f_\theta:\mbox{dom}(X) \to \mbox{dom}(Y)$ with parameters $\theta$ learned from training set  $\Dd{tr}=\{(x_0^{tr},y_0^{tr})\ldots, (x_n^{tr},y_n^{tr})\}$, we say that $\Dd{new}=\{x_0^{new},\ldots, x_k^{new}\}$ has \emph{out-of-distribution explanations} with respect to the model $f_\theta$ if $\Ss(f_\theta,\Dd{new})$ is sampled from a different distribution than $\Ss(f_\theta,\Dd{tr}_X)$.
\end{definition}

\begin{definition}[Explanation Shift]
Given a measure of statistical distance $d$,
we measure \emph{explanation shift} as the distance between two explanations of the model $f_\theta$ 
by  $d(\mathcal{S}(f_\theta, \Dd{tr}_X),\mathcal{S}(f_\theta, \Dd{new}))$.
\end{definition}

%\steffen{should explanation shift be defined on distributions or on data instances?}\carlos{since ood data is on dist, then ood exp is on dist too}

\subsection{Explanation Shift Detector: Quantifying and interpreting OOD explanations}\label{sec:Detector}

%We start by fitting a model $f_{\theta}$ to the training data, $X^{\text{tr}}$ to predict an  outcome $Y^{\text{tr}}$, and calculate the SHAP values on a hold out set $\Ss(f_\theta,X^{\text{val}})$ and in the out-of-distribution data $\Ss(f_\theta,X^{\text{ood}})$.
%\steffen{problem definitions should always be independent of the solution. Your problem definition (which should be placed at the beginning of section 3) should not even mention explanation. What should be described here is the method, i.e. that g is learned. Note that this section still misses the explanation part}\carlos{there is a paragraph there, but dont know exactly how to formalize this}\klaus{Two reasons: 1. as Steffen said: don't mix the solution into the problem. The problem here is to detect OOD Explanations. Whether you do it with a classifier $g_\psi$ or with another approach is not part of the problem. 2. I wonder if detecting ood explanations is not itself part of the solution and detecting ood data / predictions is the real problem.}\carlos{Now?}

Given a training dataset, a model, and a new dataset sampled from an unknown distribution.  The problem we are trying to solve is measuring and inspecting out-of-distribution explanations on this new dataset. Our proposed method is the \enquote{Explanation Shift Detector}:

\begin{definition}[Explanation Shift Detector]
Given training data $\Dd{tr}=\{(x_0^{tr},y_0^{tr})\ldots, (x_n^{tr},y_n^{tr})\} \sim \mathbf{P}({X} \times {Y})$  and a classifier $f_\theta$ the \emph{Explanation shift detector} returns ID, if $S(f_\theta,X)$ and
$S(f_\theta,X^{new})$ are sampled from the same distribution and OOD otherwise.
%
%aim of \enquote{Explanation OOD detection} is to train a classifier $g_\psi$ such that for any explanation of test data $\Ss(f_\theta,x)$ drawn from the mixed marginal distribution of $\Dd{te}$ and $\Dd{ood}$,$\quad g_\psi$ can correctly classify $\Ss(f_\theta,x)$ to be a realization of $\Ss(f_\theta,X)$ if ID model behaviour or $\Ss(f_\theta,X')$ if OOD model behaviour, we denote this label as $A = \{a_x | x\in X\cup X^{new}, a_x\in \{0,1\}$. The function $g:\Ss(f_\theta,\mathcal{X})\rightarrow \{ID,OOD\}$.
\end{definition}

To implement this approach we start by fitting a model $f_\theta$ to the training data, $X^{\text{tr}}$ to predict an  outcome $Y^{\text{tr}}$, and compute explanations on an in-distribution validation data set $X^{\text{val}}$:  $\Ss(f_\theta,X^{\text{val}})$. We also compute explanations on $X^{new}$, which may be in-distribution or out-of-distribution. We construct a new dataset $E=\{(S(f_\theta,x),a_x)|x\in X^{\text{val}},a_x=0\}\cup \{(S(f_\theta ,x),a_x)|x\in X^{\text{new}},a_x=1\}$ and  we train a discrimination model $g_\psi$  on the explanation space of the two distributions $E$, to predict if a certain explanation should be classified as ID or OOD.  
If the discriminator $g_\psi$  cannot distinguish the two distributions in $E$, i.e.\ its AUC is approximately $0.5$, then $X^{new}$ as a whole is classified as showing in-distribution behavior: its features interact with $f_\theta$ in the same way as with validation data. 

%\steffen{I rewrote this part. check it}

\begin{gather}\label{eq:explanationShift}
\psi = \argmin_{\tilde{\psi}} \sum_{x\in X^{\text{val}}\cup X^{\text{new}}} \ell( g_{\tilde{\psi}}(\Ss(f_\theta,x)) , 
a_x )
\end{gather}

Where $\ell$ is any given classification loss function (eq.~\ref{eq:explanationShift}). We call the model $g_\psi$ \enquote{Explanation Shift Detector}. One of the benefits of this approach is that it allows for \enquote{explaining} the \enquote{Explanation Shift Detector} at both global and individual instances levels. In this work, we use feature attribution explanations for the model $g_\psi$, whose intuition is to respond to the question of: \textit{What are the features driving the OOD model behaviour?}. For conceptual simplicity, in this work, our model $g_\psi$ is restricted to linear models and  we will use the coefficients as feature attribution methods. Future work can be envisioned in this section by applying different explainable AI techniques to the \enquote{Explanation Shift Detector}.

%Some of the main questions that arise from the previous section are \textit{What happens with multidimensional statistical testing?} or \textit{Does the explanation space work in hybrid/simultaneous shifts?}, and further more \textit{Can we interpret the feature and features interaction that drive distribution shift?}

\begin{figure}[ht]
    \centering
    \includegraphics[width=0.8\columnwidth]{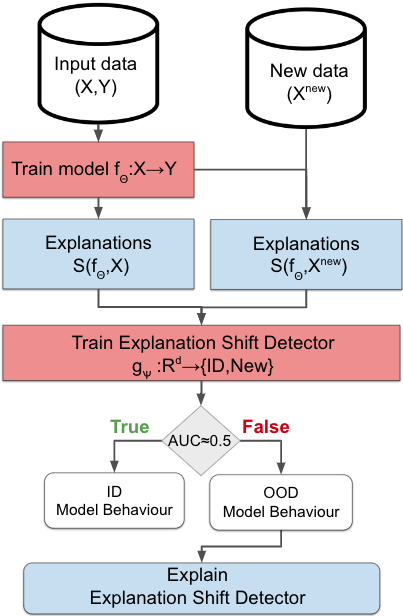}
    \caption{\enquote{Explanation Shift Detector} workflow diagram. The model $f_\theta$ is trained on $(X,Y)$, and outputs explanations for both  distributions. Then the \enquote{Explanation Shift detector} $g_\psi$ receives the explanations and aims to predict if the model behaviour is different between them. If the $AUC \neq 0.5$ then there is a change in the data that implies a change in the model behaviour. Finally, we can proceed to account $g_\psi$ with explainable AI techniques and identify the roots of the change.}
    \label{fig:workflow}
\end{figure}

\section{Mathematical Analysis}
In this section, we provide a mathematical analysis on how changes in the distributions impact the explanation space and compare to the input data space and the prediction space.

\subsection{Explanation vs Prediction spaces}\label{sec:exp.vs.pred}

%\steffen{I do not understand the following sentence. I do not even understand the preceding sentence.}\carlos{check}
This section exploits the benefits of the explanation space with respect to the prediction space when detecting distribution shifts. We will provide a theorem proving that OOD predictions imply OOD Explanations, and a counterexample for the opposite case, OOD Explanations do not necessarily imply OOD predictions. 
\begin{prop}
    Given a model $f_\theta:X \to Y$. If $f_\theta(x^{'})\neq f_\theta(x)$, then $\Ss(f_\theta,x^{'}) \neq \Ss(f_\theta,x)$.
\end{prop}
\begin{gather}
\texttt{Given}\quad f_\theta(x)\neq f_\theta(x')\\
\sum_{j=1}^p \Ss_j(f_\theta,x) = f_\theta(x) - E_X[f_\theta(X)]\\
\texttt{then}\quad \Ss(f,x)\ \neq \Ss(f,x')
\end{gather}

By additivity/efficiency property of the Shapley values~\cite{DBLP:journals/ai/AasJL21} (equation (3)), if  the prediction between two instances is different, then they differ in at least one component of their explanation vectors. 

%both terms of the equation, if the distribution of the prediction space changes, then at least one of the values of the sum has to be different, so the explanation is distinct. \gjergji{Not quite sure what you mean, because two summands can change in such a way that the sum remains unchanged and (8) could still be equality, although I understand that because of additivity/efficiency of Shapley values (8) cannot be an equality.}\carlos{If this happens the explanation space still will be different $\Ss(f,X)\neq \Ss(f,X')$}

The opposite direction does not hold, we can have out-of-distributions explanations but in-distribution predictions, as we can see in the following counter-example:
\begin{example}[Explanation shift that does not affect the prediction distribution] Given $\mathcal{D}^{tr}$ is generated from $(X_1,X_2,Y), X_1 \sim U(0,1), X_2 \sim U(1,2), Y = X_1+X_2+\epsilon$ and thus the model is $f(x)=x_1+x_2$. If $\mathcal{D}^{new}$ is generated from $X_1^{new}\sim U(1,2), X_2^{new}\sim U(0,1)$, the prediction distributions are identical $f_\theta(\mathcal{D}^{tr}),f_\theta(\mathcal{D}^{new})\sim U(0,3)$, but explanation distributions are different $S(f_\theta,\mathcal{D}^{tr}_X)\neq S(f_\theta,\mathcal{D}^{new})$
\begin{gather}
    \forall i \in  \{1,2\} \quad \Ss_i(f_\theta,x) = \alpha_i \cdot x_i  \\
   \forall i \in  \{1,2\} \Rightarrow  \Ss_i(f_\theta,X))\neq \Ss_i(f_\theta,X^{new})\\
    \Rightarrow f_\theta(X)=f_\theta(X^{new})
\end{gather}
\end{example}

%\klaus{Equation \eqref{eq:dummy_label_1}: on the left there is a vector for each sample (i.e. a matrix in total). On the right, there is a scalar for each sample (i.e. a vector in total). The equality does not make sense if the dimensions don't match!}\carlos{check?}
In this example, we can calculate the IID linear interventional Shapley values~\cite{DBLP:journals/ai/AasJL21}. Then the Shapley values for each feature will have different distributions between train and out-of-distribution explanations, but the prediction space will remain unaltered.

%We call this the Yule's effect~\cite{yule1900vii,simpson1951interpretation} of distribution shift on the predictions, where the differences between feature contribution can cancel out between themselves and result in an equal prediction, even though the model behaviour is distinct. \carlos{Klaus, can you check if this makes sense?}

%\steffen{sometimes you say `output space', here you say `prediction space'. I get confused here. Also the but sentence is unclear. Same as what?}\carlos{Let's use \enquote{Prediction Space} all the time.}

\subsection{Explanation shifts vs input data distribution shifts}\label{subsec:explanationShiftMethods}

This section provides two different examples of situations where changes in the explanation space can correctly account for model behavior changes, but where statistical checks on the input data either (1) cannot detect changes, 
%(2) require sophisticated methods to detect these changes,
or (2) detect changes that do not affect model behavior. For simplicity the model used  in the analytical examples is a linear regression where, if the features are independent, the Shapley value can be estimated by $\Ss(f_\theta, x_i) = a_i(x_i-\mu_i)$, where $a_i$ are the coefficients of the linear model and $\mu_i$ the mean of the features \cite{DBLP:journals/corr/ShapTrueModelTrueData}. 

\subsubsection{Detecting multivariate shift}
One  type of distribution shift that is challenging to detect comprises cases where the univariate distributions for each feature $j$ are equal between the source $\Dd{tr}$ and the unseen dataset $\Dd{new}$, but where  interdependencies among different features change. 
%\steffen{I do not understand the following sentence. You say in the following sentence that univariate tests are sufficient, though in the preceding sentence  you basically said that they are not sufficient, and then you say that your univariate distribution is on high dimensional space, which brings another aspect in the discussion. This later aspect needs to be shaved off into a different sentence and anyway there is not always a high dimensional space.}\carlos{check?}
On the other hand, multi-covariance statistical testing is not an easy task that has high sensitivity easily leading to false positives. The following  examples demonstrate that Shapley values account for co-variate interaction changes while a univariate statistical test will provide false negatives. \\
\textbf{Example 1: \textit{Multivariate Shift}}\\ 
\textit{Let $X = (X_1,X_2) \sim {\tiny  N\left(\begin{bmatrix}\mu_{1}  \\ \mu_{2} \end{bmatrix},\begin{bmatrix}\sigma^2_{x_1} & 0 \\0 & \sigma^2_{x_2} \end{bmatrix}\right),}$\\
 $X^{new} = (X^{new}_1,X^{new}_2) \sim {\tiny N\left(\begin{bmatrix}\mu_{1}  \\ \mu_{2} \end{bmatrix},\begin{bmatrix} \sigma^2_{x_1} & \rho\sigma_{x_1}\sigma_{x_2}  \\ \rho\sigma_{x_1}\sigma_{x_2} & \sigma^2_{x_2}\end{bmatrix}\right)}$. We fit a linear model 
$f_\theta(X_1,X_2) = \gamma + a\cdot X_1 + b \cdot X_2.\hspace{0.5cm}$  $X_1$ and $X_2$ are identically distributed with $X_1^{new}$ and $X_2^{new}$, respectively, while this does not hold for the corresponding SHAP values $\Ss_j(f_\theta,X)$ and $\Ss_j(f_\theta,X^{new})$.}

The  detailed analysis is given in the Appendix.

%The above theorem works under the assumption of linear regression and that the covariate term $\rho=1$, is a not-so-common situation in ML applications.
\begin{comment}
\subsubsection{Detecting concept distribution shift}
One  type of distribution shift that is most challenging to detect comprises cases where distributions are equal between source and unseen data-set $P(X^{tr}) = P(X^{new})$ and the target variable  $P(Y^{tr}) = P(Y^{new})$ and what changes are the relationships that features have with the target $P(Y^{tr}|X^{tr}) \neq  P(Y^{new}|X^{new})$. This kind of distribution shift is also known as concept drift or posterior shift~\cite{DesignMLSystems} and is especially difficult to notice, as it requires labeled data to detect in general. The following example compares how  explanations may change for two models fed with the same input data and different target relations.

\textbf{Example 2: \textit{Concept shift}}\textit{
Let $X = (X_1,X_2) \sim N(\mu,I)$, and $X^{new}= (X^{new}_1,X^{new}_2) \sim N(\mu,I)$, where $I$ is an identity matrix of order two and $\mu = (\mu_1,\mu_2)$. We now create two synthetic targets $Y=a + \alpha \cdot X_1 + \beta \cdot X_2 + \epsilon$ and $Y^{new}=a + \beta \cdot X_1 + \alpha \cdot X_2 + \epsilon$. Let $f_\theta$ be a linear regression model trained on $(X,Y)$ and $h_\phi$ another linear model trained on $(X^{new},Y^{new})$. Then $P(f_\theta(X)) = P(h_\phi(X^{new}))$, $P(X) = P(X^{new})$ but $\Ss(f_\theta,X)\neq \Ss(h_\phi, X)$}. 
\end{comment}
\subsubsection{Shifts on uninformative features by the model}

Another typical problem is false positives when a statistical test recognizes differences between a source distribution and a new distribution, though the differences do not affect the model behavior\cite{grinsztajn2022why}. One of the intrinsic properties that Shapley values satisfy is the \enquote{Dummy}, where a feature $j$ that does not change the predicted value, regardless of which coalition the feature is added, should have a Shapley value of $0$. If $\mathrm{val}(S\cup \{j\}) = \mathrm{val}(S)$ for all $S\subseteq \{1,\ldots, p\}$ then $\Ss_j(f_\theta, x_i)=0$.

\textbf{Example 2: \textit{Unused features}}\textit{
Let $X = (X_1,X_2,X_3) \sim N(\mu,\mathrm{diag}(c))$, and $X^{new}= (X^{new}_1,X^{new}_2,X^{new}_3) \sim N(\mu,\mathrm{diag}(c'))$, where $\mathrm{c}$ and $\mathrm{c'}$ are an identity matrix of order three and $\mu = (\mu_1,\mu_2,\mu_3)$. We now create a synthetic target $Y=a_0 + a_1 \cdot X_1 + a_2 \cdot X_2 + \epsilon$ that is independent of $X_3$. We train a linear regression $f_\theta$ on $(X,Y)$, with coefficients $a_0,a_1,a_2,a_3$. Then if $\mu_3'\neq \mu_3$ or $c_3' \neq c_3$, then $P(X_3)$ can be different from $P(X_3^{new})$ but $\Ss_3(f_\theta, X) = \Ss_3(f_\theta,X^{new})$}
%\klaus{If $c$ and $c'$ are vectors, then $c\cdot I$ is also a vector, but you need a covariance \textbf{matrics}. Write $\mathrm{diag}(c)$ instead.}\carlos{does not it seem weird?}\klaus{We can make it less weird by (a) using \texttt{\textbackslash mathrm} instead of \texttt{\textbackslash texttt} and (b) omitting $\cdot I$: $\mathrm{diag}(c)$ is already a matrics}

\section{Experiments}\label{sec:experiments}

The first experimental section explores the detection of distribution shift on the previoust synthetic examples. The second part explores the utility of explanation shift on real data applications.

\subsection{Explanation Shift Detection}
\subsubsection{Detecting multivariate shift}\label{sec:multivariate}

Given two bivariate normal distributions $X = (X_1,X_2) \sim  N\left(0,\begin{bmatrix}1 & 0 \\0& 1 \end{bmatrix}\right)$ and $X^{new} = (X^{new}_1,X^{new}_2) \sim  N \left( 0,\begin{bmatrix}1 & 0.2 \\0.2 & 1 \end{bmatrix}\right)$, then, for each feature $j$ the underlying distribution is equally distributed between $X$ and $X^{new}$, $\forall j \in \{1,2\}: P(X_j) = P(X^{new}_j)$, and what is different are the interaction terms between them. We now create a synthetic target $Y=X_1\cdot X_2 + \epsilon$ with $\epsilon \sim N(0,0.1)$ and fit a gradient boosting decision tree  $f_\theta(X)$. Then we compute the SHAP explanation values for $\mathcal{S}(f_\theta,X)$ and $\mathcal{S}(f_\theta,X^{new})$

\begin{table}[ht]
\centering
\caption{Displayed results are the one-tailed p-values of the Kolmogorov-Smirnov test comparison between two underlying distributions. Small p-values indicates that compared distributions would be very unlikely  to be equally distributed. SHAP values correctly indicate the interaction changes that individual distribution comparisons cannot detect}\label{table:multivariate}
\begin{tabular}{c|cc}
Comparison                                 & \textbf{p-value} & \textbf{Conclusions} \\ \hline
$P(X_1)$, $P(X^{new}_1)$                        & 0.33                        & Not Distinct                         \\
$P(X_2)$, $P(X^{new}_2)$                        & 0.60                        & Not Distinct                          \\
$\Ss_1(f_\theta,X)$, $\Ss_1(f_\theta,X^{new})$ & $3.9\mathrm{e}{-153}$        & Distinct                              \\
$\Ss_2(f_\theta,X)$, $\Ss_2(f_\theta,X^{new})$ & $2.9\mathrm{e}{-148}$        & Distinct   
\end{tabular}
\end{table}

Having drawn $50,000$ samples from both $X$ and $X^{new}$, in Table~\ref{table:multivariate}, we evaluate whether changes on the input data distribution or on the explanations are able to detect changes on covariate distribution.
%\steffen{... complete... What are your hypotheses/null-hypotheses? You jump over too many steps, which makes your proposition become awkwardly ambiguous.}
For this, we compare the one-tailed p-values of the Kolmogorov-Smirnov test between the input data distribution, and the explanations space.  Explanation shift correctly detects the multivariate distribution change that univariate statistical testing can not detect.

\subsubsection{Uninformative features on synthetic data}

To have an applied use case of the synthetic example from the methodology section, we create a three-variate normal distribution $X = (X_1,X_2,X_3) \sim N(0,I_3)$, where $I_3$ is an identity matrix of order three. The target variable is generated  $Y=X_1\cdot X_2 + \epsilon$ being independent of $X_3$. For both, training and test data, $50,000$ samples are drawn. Then out-of-distribution data is created by shifting $X_3$, which is independent of the target, on test data $X^{new}_3= X^{te}_3+1$.

\begin{table}[ht]
\centering
\caption{Distribution comparison when modifying a random noise variable on test data. The input data shifts while explanations and predictions do not.}\label{table:unused}
\begin{tabular}{c|c}
Comparison                                              & \textbf{Conclusions} \\ \hline
$P(X^{te}_3)$, $P(X^{new}_3)$                                       & Distinct                \\
$f_\theta(X^{te})$, $f_\theta(X^{new})$                     & Not Distinct            \\
$\Ss(f_\theta,X^{te})$, $\Ss(f_\theta,X^{new})$                    & Not Distinct            \\
\end{tabular}
\end{table}

In Table~\ref{table:unused}, we see how an unused feature has changed the input distribution, but the explanation space and performance evaluation metrics remain the same. The \enquote{Distinct/Not Distinct} conclusion is based on the one-tailed p-value of the Kolmogorov-Smirnov test drawn out of $50,000$ samples for both distributions.

\subsubsection{Explanation shift that does not affect the prediction}

In this case we provide a situation when we have changes in the input data distributions that affect the model explanations but do not affect the model predictions due to positive and negative associations between the model predictions and the distributions cancel out producing a vanishing correlation in the mixture of the distribution (Yule's effect~\ref{sec:exp.vs.pred}).  

We create a train and test data by drawing $50,000$ samples from a bi-uniform distribution  $X_1 \sim U(0,1), \quad X_2 \sim U(1,2)$ the target variable is generated  by $Y = X_1+X_2$ where we train our model $f_\theta$. Then if out-of-distribution data is sampled from $X_1^{new}\sim U(1,2)$, $X_2^{new}\sim U(0,1)$

\begin{table}[ht]
\centering
\caption{Distribution comparison over how the change on the contributions of each feature can cancel out to produce an equal prediction (cf. Section \ref{sec:exp.vs.pred}), while explanation shift will detect this behaviour changes on the predictions will not.}\label{table:predShift}
\begin{tabular}{c|c}
Comparison                                              & \textbf{Conclusions} \\ \hline
$f(X^{te})$, $f(X^{new})$                                  & Not Distinct            \\
$\Ss(f_\theta,X^{te}_2)$, $\Ss(f_\theta,X^{new}_2)$                    & Distinct            \\
$\Ss(f_\theta,X^{te}_1)$, $\Ss(f_\theta,X^{new}_1)$                    & Distinct            \\
\end{tabular}
\end{table}

In Table~\ref{table:predShift}, we see how an unused feature has changed the input distribution, but the explanation space and performance evaluation metrics remain the same. The \enquote{Distinct/Not Distinct} conclusion is based on the one-tailed p-value of the Kolmogorov-Smirnov test drawn out of $50,000$ samples for both distributions.

\subsection{Explanation Shift Detector: Measurements on synthetic data}

In this work, we are proposing explanation shifts as an indicator of out-of-distribution model behavior. The \textit{Explanation Shift Detector} (cf. Section \ref{sec:Detector}), aims to detect if the behaviour of a machine learning model is different between unseen data and training data.

As ablation studies, in this work, we compare our method that learns on the explanations space (eq. \ref{eq:explanationShift}) against learning on different spaces: on input data space, that detects out-of-distribution data, but is independent of the model:
%\gjergji{The two equations are not well explained. What is $n$? How do you know that argmin of the average loss has a unique solution $\psi$? By the way, I see the same problems in Equation (4).}\carlos{We discused this with Klaus. The idea is that  argmin refers to the optimal paramer/arg. When do I say that it has an unique solution?} \gjergji{Note that in general argmin(...) can have multiple solutions (in this case multiple optimal parameters), unless you can prove some kind of convexity; it is better to write just min(...) instead (i.e., not an equation) and say that we are interested in parameters $\psi$ that minimize ... }
%\steffen{I think that it is not a good idea to give these different functions the same name. In Table 5 it becomes visible that different functions are learned in (11), in (12) and in definition 3.6, but this is far from clear when reading this text. If different names were given, they would be clearly distinguished in Table 5.}\carlos{I have used different params}

\begin{gather}
\phi = \argmin_{\tilde{\phi}} \sum_{x\in X\cup X^{new}} \ell( g_{\tilde{\phi}}(\textcolor{blue}{x})), a_x )
\end{gather}
on output space, that detects out-of-distribution predictions, but can suffer from Yule's effect of distribution shift (cf. section~\ref{sec:exp.vs.pred}):
\begin{gather}
\Upsilon = \argmin_{\tilde{\Upsilon}} \sum_{x\in X\cup X^{new}} \ell( g_{\tilde{\Upsilon}}(\textcolor{blue}{f_\theta(x)}), a_x )
\end{gather}

Our first experiment showcases the two main contributions of our method: $(i)$ being more sensitive than output spaces and input spaces to changes in the model behaviour and, $(ii)$ accounting for its drivers. 

In order to do this, we first generate a synthetic dataset with a shift similar to the multivariate shift one (cf. Section \ref{sec:multivariate}), but we add an extra variable $X_3 = N(0,1)$ and generate our target $Y=X_1 \cdot X_2 + X_3$, and parametrize the multivariate shift between $\rho = r(X_1,X_2)$. For the model $f_\theta$ we use a gradient boosting decision tree, while for $g_\psi$ we use a logistic regression. For model performance metrics, as we are in binary classification scenarios, we use the Area Under the Curve (AUC)---a  metric that illustrates the diagnostic ability of a binary classifier system as its discrimination threshold is varied~\cite{statisticallearning}.

\begin{table}[ht]
\caption{Conceptual comparison table over different detection methods over the examples discussed above. The \enquote{Explanation Shift Detector}, learning $g_\psi$ over the explanation space is the only method that achieves the desired results and is accountable. We evaluate accountability by checking if the feature attributions of the detection method correspond with the synthetic shift generated in both scenarios}\label{tab:ExplanationShiftDetector}
\begin{tabular}{c|cccc}
\textbf{Detection Method} & \textbf{Multiv.}        & \textbf{Uninf.} & \textbf{Accountability} \\ \hline
Explanation sp. ($g_\psi$)& \cmark & \cmark & \cmark \\
Input space($g_\phi$)     & \cmark & \xmark  & \xmark \\
Prediction sp.($g_\Upsilon$)  & \cmark & \cmark &  \xmark \\
Input Shift(Univ)         & \xmark & \xmark & \xmark \\
Input Shift(Mult.)        & \cmark & \xmark & \xmark \\
Output Shift              & \cmark & \cmark & \xmark \\
Uncertainty           & $\sim$ & \cmark & \cmark 
\end{tabular}
\end{table}

\begin{figure*}[ht]
\centering
\includegraphics[width=.49\textwidth]{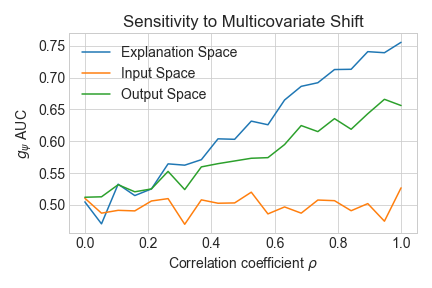}\hfill
\includegraphics[width=.49\textwidth]{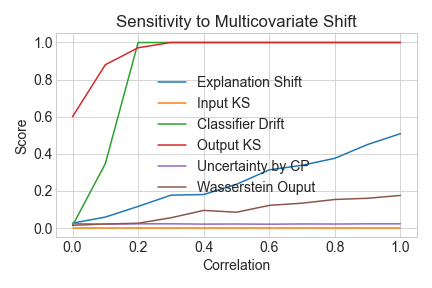}
\caption{In the left figure, comparison of the performance of \textit{Explanation Shift Detector}, based on learning on  different spaces. Learning on the explanation space shows more sensitivity than learning in the output space. In the right figure, related work comparison of SOTA methods \cite{alibi-detect,mougan2022monitoring,DBLP:conf/nips/RabanserGL19}, good indicators should follow a progressive steady positive slope.}
\label{fig:sensitivity}
\end{figure*}

In Table \ref{tab:ExplanationShiftDetector} and Figure \ref{fig:sensitivity}, we show the effectos of our algorithmic approach when learning on different spaces. In the sensitivity experiment, we see that the Explanation Space offers a higher sensitivity towards distribution shift detection. This can be explained using the additive property of the Shapley values. What the explanation space is representing is a higher dimensional space than the output space that contains the model behavior. On the other hand, the input space, even if it has the same dimensional, it does not contain the projection of the model behaviour. Furthermore, we can proceed and explain what are the reasons driving the drift, by extracting the coefficients of $g_\psi$ 
 of the $\rho = 1$ case, $\beta_1 \neq 0, \beta_2\neq 0, \beta_3\sim 0 $, providing global explainability about the features that are shifting, the \textit{Explanation Shift Detector} correctly detects the features that are causing the drift.

In the right image of Figure \ref{fig:sensitivity} the comparison\footnote{The metric for the \enquote{Explanation Shift Detector} is $2\cdot(AUC-0.5)$, in order to scale respect to other metrics.} against other state-of-the-art techniques: statistical differences on the input data (Input KS, Classifier Drift)~\cite{alibi-detect,continual_learning}, that are independent of the model; uncertainty estimation ~\cite{DBLP:conf/nips/KimXB20,mougan2022monitoring,romano2021pmlb}, whose properties under specific types of shift remains unknown, or statistical changes on the output data~\cite{fort2021exploring,NEURIPS2020_219e0524} (K-S and Wasserstein Distance), which correctly detect that the model behaviour is changing, but lacks the sensitivity of the explanation space.

%has, and measures on p-values are over-saturated. \gjergji{What do you mean by this? How can a measure be over-saturated?} %In the Appendix (cf. Section XX) we provide other synthetic data experiment with Unused Features. \carlos{Yet to do.}

\subsection{Experiments on real data: Inspecting out-of-distribution explanations}
\begin{figure*}[ht]
\centering
\includegraphics[width=.49\textwidth]{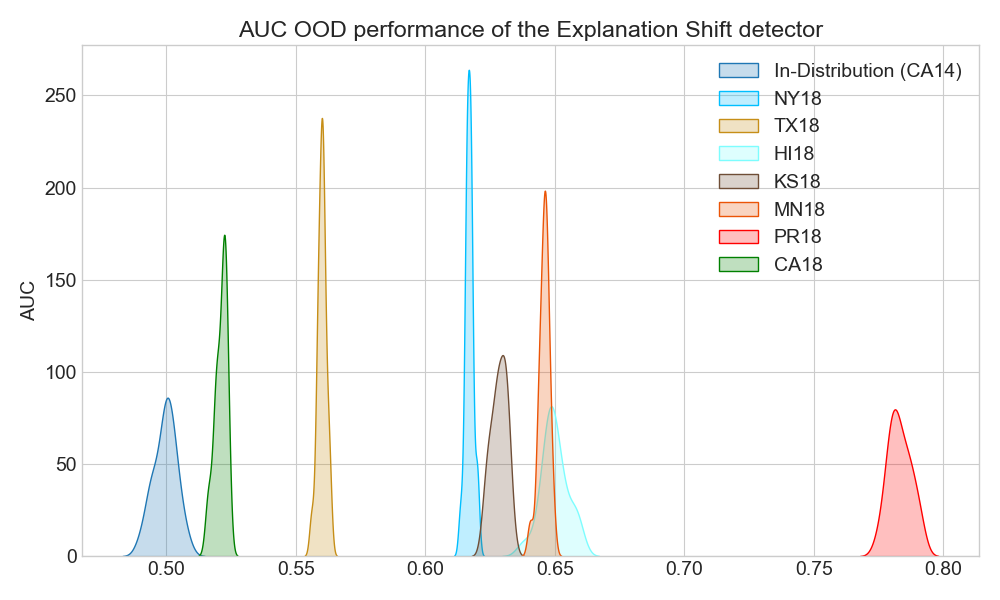}\hfill
\includegraphics[width=.49\textwidth]{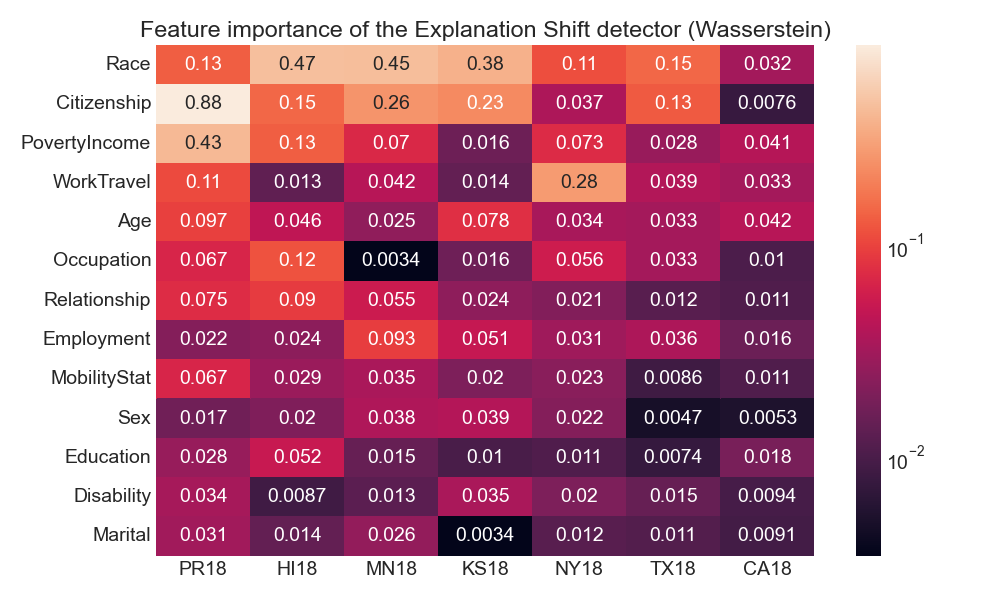}
\caption{In the left figure, comparison of the performance of \textit{Explanation Shift Detector}, in different states. Learning in the explanation space shows more sensitivity than learning in the output space. In the right figure, strength analysis of features driving the change in model behaviour.}
\label{fig:xai}
\end{figure*}

The need for explainable model monitoring has been expressed by several authors~\cite{https://doi.org/10.48550/arxiv.2007.06299,continual_learning,desiderataECB,DBLP:conf/aistats/BudhathokiJBN21}, understanding the effects of the distribution shift on the behaviour can provide algorithmic transparency to stakeholders and to the ML engineering team.

After providing analytical examples and experiments with synthetic data, still there is the challenge of operationalizing the \textit{Explanation Shift Detector} to real-world data.  In this section, we provide experiments on the ACSTravelTime task, whose goal is to predict whether an individual has a commute to work that is longer than 20 minutes. To create the distribution shift we train the model $f_\theta$ in California in 2014 and evaluating in the rest of the states in 2018, creating a geopolitical and temporal shift. The model $g_\theta$ is trained each time on each state using only the $X^{new}$ in the absence of the label, and its performance is evaluated by a 50/50 random train-test split. As models we use a gradient boosting decision tree\cite{xgboost,catboost} as estimator $f_\theta$, approximating the Shapley values by TreeExplainer \cite{lundberg2020local2global}, and using logistic regression for the \textit{Explanation Shift Detector}.

Our hypothesis is that the AUC in OOD data of the \enquote{Explanation Shift Detector} $g_\psi$ will be higher than ID data due to OOD model behaviour. In Figure \ref{fig:xai}, we can see the performance of the \enquote{Explanation Shift Detector} over different 
 data distributions. The baseline is a hold-out set of $ID-CA14$, and the closest AUC is for $CA18$, where there is just a temporal shift, then the OOD detection performance over the rest of states. The most OOD state is Puerto Rico (PR18).

The next question that we aim to answer is what are the features where the model behaviour is different. For this, we do a distribution comparison between the linear coefficients of the \enquote{Explanation Shift Detector} in ID and in OOD. As a distance measure, we use the Wasserstein distance between 1000 in-distribution bootstraps using a $63.2\%$ sampling fraction~\cite{statisticallearning} from California-14 and 1000 OOD bootstraps from other states in 2018 (see \ref{fig:xai}). In the right image  \ref{fig:xai}, for PR18, we see that the most important feature is the citizenship status\footnote{See the ACS PUMS data dictionary for the full list of variables available \url{https://www.census.gov/programs-surveys/acs/microdata/documentation.html}}. 

We also perform an across-task evaluation, by comparing it with the other prediction task in the appendix. We can see how, even if some of the features are in present in different prediction tasks, the weight and importance order assigned by the \enquote{Explanation Shift Detector} is different. One of the main contributions of this method is that is not just how distribution differs, but how they differ with respect to the model.

\section{Discussion}

In this work, we have proposed explanation shifts as a key indicator for investigating the interaction between distribution shifts and the learned model. Finding that monitoring explanations shift is a better indicator than model varying behaviour.

Our approach is not able to detect concept shifts, as concept shift requires understanding the interaction between prediction and response variables. By the very nature of concept shifts, such changes can only be understood if new data comes with labeled responses. We work under the assumption that such labels are not available for new data and, therefore, our method is not able to predict the degradation of prediction performance under distribution shifts. All papers such as \cite{garg2022leveraging,mougan2022monitoring,baek2022agreementontheline,chen2022estimating,fang2022is,baek2022agreementontheline,DBLP:conf/icml/MillerTRSKSLCS21} that address the monitoring of prediction performance have the same limitation. Only under specific assumptions, e.g., no occurrence of concept shift, performance degradation can be predicted with reasonable reliability.

%Without any assumptions on the type of shift, estimating model performance in out-of-distribution data is a challenging task, where no estimator will be the best under all the types of shift \cite{garg2022leveraging}, in this work we have  shifted the focus on detecting out-of-distribution model behaviour instead of out-of-distribution model performance or out-of-distribution data. We compared how well measures of explanation shift would perform relative to distribution shift and found encouraging results. 
The  potential utility of explanation shifts as distribution shift indicators that affect the model in computer vision or natural language processing tasks remains an open question. We have used Shapley values to derive indications of explanation shifts, but we believe that other AI explanation techniques may be applicable and come with their own advantages.

%As explainability technique we have used Shapley values, there is further explainable AI techniques such as more feature attribution methods or contrafactual explanations that can be applied. 

%We can see how for CA14 and CA18 the statistical differences are lower, indicating that the coefficients are more likely to be from the same distribution and that there are no out-of-distribution model behaviour drivers. On the Appendix (cf. Section XX) we provide experiments on 5 more datasets based on the US Income data. \carlos{Yet to do.}

%\\[5cm]
%\steffen{What I have been preaching for a while, but which you seem to ignore is:
%\\
%1. You need to compare to state-of-the-art methods. Pull out 2-3 methods published in the last 2-3 years and compare your approach.
%\\
%2. You need to go beyond the census dataset. 
%\\
%I believe that the paper will be rejected if you do not do these two steps. They seem unnecessary to you, but these are standard check boxes that reviewers want to have ticked. When they feel even slightly uneasy about your approach, e.g. because it is so simple as it is, they will give the lack of 1. and 2. as reasons that you did not do your job fully}

\section{Conclusions}
Traditionally, the problem of detecting model shift behaviour has relied on measurements for detecting shifts in the input or output data distributions. In this paper, we have provided theoretical and experimental evidence that explanation shift can be a more suitable indicator to detect and identify the shift in the behaviour of machine learning models. We have provided mathematical analysis examples, synthetic data, and real data experimental evaluation. We found that measures of explanation shift can provide more insights than measures of the input distribution and prediction shift when monitoring machine learning models.

\subsection*{Reproducibility Statement}\label{sec:reproducibility}
To ensure reproducibility, we make the data, code repositories, and experiments publicly available
\footnote{\url{https://anonymous.4open.science/r/ExplanationShift-icml/README.md}}. Also, an open source Python package is released with the methods used (to be released upon acceptance). 
For our experiments, we used default \texttt{scikit-learn} parameters \cite{pedregosa2011scikit}. We describe the system requirements and software dependencies of our experiments. Experiments were run on a 4 vCPU server with 32 GB RAM.

\begin{comment}
\subsection*{Acknowledgments}
This work has received funding by the European Union’s Horizon 2020 research and innovation programme under the Marie
Skłodowska-Curie Actions (grant agreement number 860630) for
the project : \enquote{NoBIAS - Artificial Intelligence without Bias}. Furthermore, this work reflects only the authors’ view and the European Research Executive Agency (REA) is not responsible for any
use that may be made of the information it contains. 
\end{comment}

\bibliography{references}
\bibliographystyle{icml2023}

%%%%%%%%%%%%%%%%%%%%%%%%%%%%%%%%%%%%%%%%%%%%%%%%%%%%%%%%%%%%%%%%%%%%%%%%%%%%%%%
%%%%%%%%%%%%%%%%%%%%%%%%%%%%%%%%%%%%%%%%%%%%%%%%%%%%%%%%%%%%%%%%%%%%%%%%%%%%%%%
% APPENDIX
%%%%%%%%%%%%%%%%%%%%%%%%%%%%%%%%%%%%%%%%%%%%%%%%%%%%%%%%%%%%%%%%%%%%%%%%%%%%%%%
%%%%%%%%%%%%%%%%%%%%%%%%%%%%%%%%%%%%%%%%%%%%%%%%%%%%%%%%%%%%%%%%%%%%%%%%%%%%%%%
\newpage
\onecolumn
\appendix
\section{Analytical examples}
This section covers the analytical examples demonstrations presented in the Section \ref{subsec:explanationShiftMethods} of the main body of the paper.
\subsection{Explanation vs Prediction}

\begin{prop}
    Given a model $f_\theta:X \to Y$. If $f_\theta(x^{'})\neq f_\theta(x)$, then $\Ss(f_\theta,x^{'}) \neq \Ss(f_\theta,x)$.
\end{prop}
\begin{gather}
\texttt{Given}\quad f_\theta(x)\neq f_\theta(x')\\
\sum_{j=1}^p \Ss_j(f_\theta,x) = f_\theta(x) - E_X[f_\theta(X)]\\
\texttt{then}\quad \Ss(f,x)\ \neq \Ss(f,x')
\end{gather}

\begin{example}[Explanation shift that does not affect the prediction distribution] Given $\mathcal{D}^{tr}$ is generated from $(X_1,X_2,Y), X_1 \sim U(0,1), X_2 \sim U(1,2), Y = X_1+X_2+\epsilon$ and thus the model is $f(x)=x_1+x_2$. If $\mathcal{D}^{new}$ is generated from $X_1^{new}\sim U(1,2), X_2^{new}\sim U(0,1)$, the prediction distributions are identical $f_\theta(\mathcal{D}^{tr}),f_\theta(\mathcal{D}^{new})\sim U(0,3)$, but explanation distributions are different $S(f_\theta,\mathcal{D}^{tr}_X)\neq S(f_\theta,\mathcal{D}^{new})$
\begin{gather}
    \forall i \in  \{1,2\} \quad \Ss_i(f_\theta,x) = \alpha_i \cdot x_i  \\
   \forall i \in  \{1,2\} \Rightarrow  \Ss_i(f_\theta,X))\neq \Ss_i(f_\theta,X^{new})\\
    \Rightarrow f_\theta(X)=f_\theta(X^{new})
\end{gather}
\end{example}
\subsection{Explanation shifts vs input data distribution shifts}
\subsubsection{Multivariate shift}
\textbf{Example 1: \textit{Multivariate Shift}}\textit{
Let $X = (X_1,X_2) \sim  N\left(\begin{bmatrix}\mu_{1}  \\ \mu_{2} \end{bmatrix},\begin{bmatrix}\sigma^2_{x_1} & 0 \\0 & \sigma^2_{x_2} \end{bmatrix}\right)$
and $X^{ood} = (X^{ood}_1,X^{ood}_2) \sim  N\left(\begin{bmatrix}\mu_{1}  \\ \mu_{2} \end{bmatrix},\begin{bmatrix} \sigma^2_{x_1} & \rho\sigma_{x_1}\sigma_{x_2}  \\ \rho\sigma_{x_1}\sigma_{x_2} & \sigma^2_{x_2}\end{bmatrix}\right)$ and target $Y = X_1 + X_2 + \epsilon$. We fit a linear model 
$f_\theta(X_1,X_2) = \gamma + a\cdot X_1 + b \cdot X_2.\hspace{0.5cm}$  $X_1$ and $X_2$ are identically distributed with $X_1^{ood}$ and $X_2^{ood}$, respectively, while this does not hold for the corresponding SHAP values $\Ss_j(f_\theta,X)$ and $\Ss_j(f_\theta,X^{ood})$.}
\begin{gather}
\Ss_1(f_\theta,x) = a(x_1 - \mu_1)\\
\Ss_1(f_\theta,x^{ood}) =\\
=\frac{1}{2}[\mathrm{val}(\{1,2\}) - \mathrm{val}(\{2\})] + \frac{1}{2}[\mathrm{val}(\{1\}) - \mathrm{val}(\emptyset)] \\
\mathrm{val}(\{1,2\}) = E[f_\theta|X_1=x_1, X_2=x_2] = a x_1 + b x_2\\
\mathrm{val}(\emptyset) = E[f_\theta]= a \mu_1 + b  \mu_2 \\
\mathrm{val}(\{1\}) = E[f_\theta(x) | X_1 = x_1] +b\mu_2 \\
\mathrm{val}(\{1\}) = \mu_1 +\rho \frac{\rho_{x_1}}{\sigma_{x_2}}(x_1-\sigma_1)+b \mu_2\\
\mathrm{val}(\{2\}) = \mu_2 +\rho \frac{\sigma_{x_2}}{\sigma_{x_1}}(x_2-\mu_2)+a\mu_1 \\
\Rightarrow \Ss_1(f_\theta,x^{ood})\neq a(x_1 - \mu_1)
\end{gather}
\subsubsection{Uninformative Features}

\textbf{Example 2: \textit{Unused features}}\textit{
Let $X = (X_1,X_2,X_3) \sim N(\mu,\mathrm{diag}(c))$, and $X^{ood}= (X^{ood}_1,X^{ood}_2,X^{ood}_3) \sim N(\mu,\mathrm{diag}(c'))$, where $\mathrm{c}$ and $\mathrm{c'}$ are an identity matrix of order three and $\mu = (\mu_1,\mu_2,\mu_3)$. We now create a synthetic target $Y=a_0 + a_1 \cdot X_1 + a_2 \cdot X_2 + \epsilon$ that is independent of $X_3$. We train a linear regression $f_\theta$ on $(X,Y)$, with coefficients $a_0,a_1,a_2,a_3$. Then if $\mu_3'\neq \mu_3$ or $c_3' \neq c_3$, then $P(X_3)$ can be different from $P(X_3^{ood})$ but $\Ss_3(f_\theta, X) = \Ss_3(f_\theta,X^{ood})$}
\begin{gather}
X_3\sim N(\mu_3,c_3),X_3^{ood} \sim N(\mu_3^{'}, c_3^{'})\\
\mathrm{If} \quad  \mu_3^{'}\neq \mu_3 \quad\mathrm{or} \quad c_3^{'}\neq c_3 \rightarrow P(X_3)\neq P(X_3^{ood})\\
\Ss(f_\theta,X) = \left(\begin{bmatrix} a_1(X_1 - \mu_1)  \\a_2(X_2 - \mu_2)  \\a_3(X_3 - \mu_3)   \end{bmatrix} \right) = \left(\begin{bmatrix} a_1(X_1 - \mu_1)  \\a_2(X_2 - \mu_2)  \\0   \end{bmatrix} \right)\\
\Ss_3(f_\theta,X)= \Ss_3(f_\theta, X^{ood})
\end{gather}

%\section{Synthetic data experiments}
%This section covers the last experiment of uninformative features on synthetic data that aims at providing empirical evidence about using the explanation space as (cf. Section \ref{sec:experiments})  

\section{Experiments on real data}
In this section, we extend the prediction task of the main body of the paper.  The methodology used follows the same structure, we start by creating a distribution shift by training the model $f_\theta$ in California in 2014 and evaluating it in the rest of the states in 2018, creating a geopolitical and temporal shift. The model $g_\theta$ is trained each time on each state using only the $X^{ood}$ in the absence of the label, and its performance is evaluated by a 50/50 random train-test split. As models we use a gradient boosting decision tree\cite{xgboost,catboost} as estimator $f_\theta$, approximating the Shapley values by TreeExplainer \cite{lundberg2020local2global}, and using logistic regression for the \textit{Explanation Shift Detector}.

\subsection{ACS Employment}
The goal of this task is to predict whether an individual, between the ages of 16 and 90, is employed. For this prediction task the AUC of all the other states (except PR18) falls below $0.60$, indicating not high OOD explanations. For the most OOD state, PR18, the \enquote{Explanation Shift Detector} finds that the model has shifted due to features such as Citizenship or Military Service.

\begin{figure*}[ht]
\centering
\includegraphics[width=.49\textwidth]{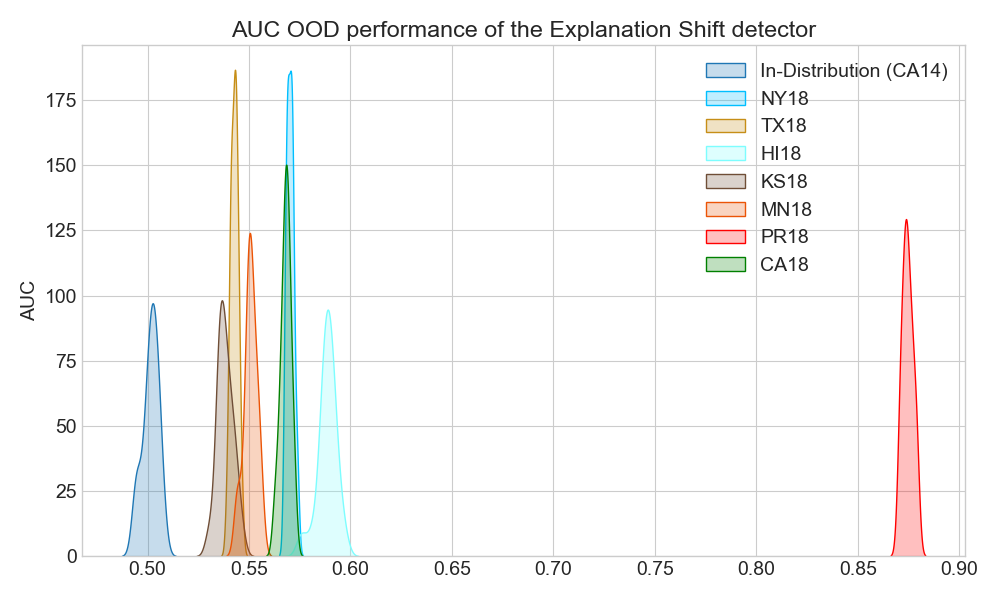}\hfill
\includegraphics[width=.49\textwidth]{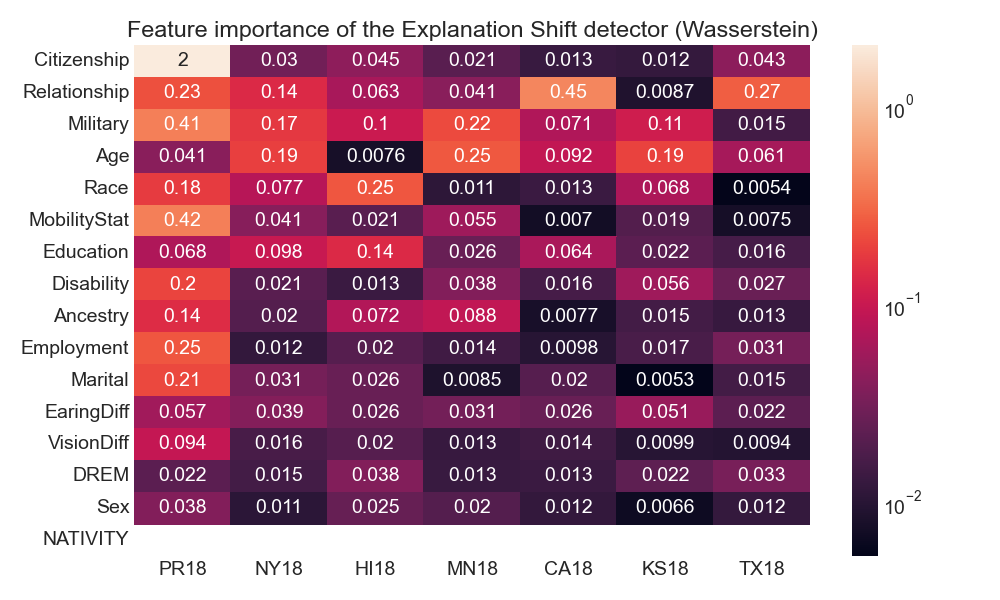}
\caption{In the left figure, comparison of the performance of \textit{Explanation Shift Detector}, in different states for the ACS Employment dataset. For this dataset, most of the states have the same OOD detection AUC, except for PR18. This difference in the model behaviour is due to features such as Citizenship and Military Service. Features such as difficulties in hearing or seeing, do not play a role in the OOD model behaviour.}
\label{fig:xai.employment}
\end{figure*}

\subsection{ACS Income}
 The goal of this task is to predict whether an individual's income is above $\$50,000$, only includes individuals above the age of 16, and report an income of at least $\$100$. This dataset can serve as a comparable replacement to the UCI Adult dataset.

 For this prediction task the results are different from the previous two cases, the state with the highest OOD score is $KS18$, with the \enquote{Explanation Shift Detector} highlighting features as Place of Birth, Race or Working Hours Per Week. The closest state to ID is CA18, where there is only a temporal shift without any geospatial distribution shift.
\begin{figure*}[ht]
\centering
\includegraphics[width=.49\textwidth]{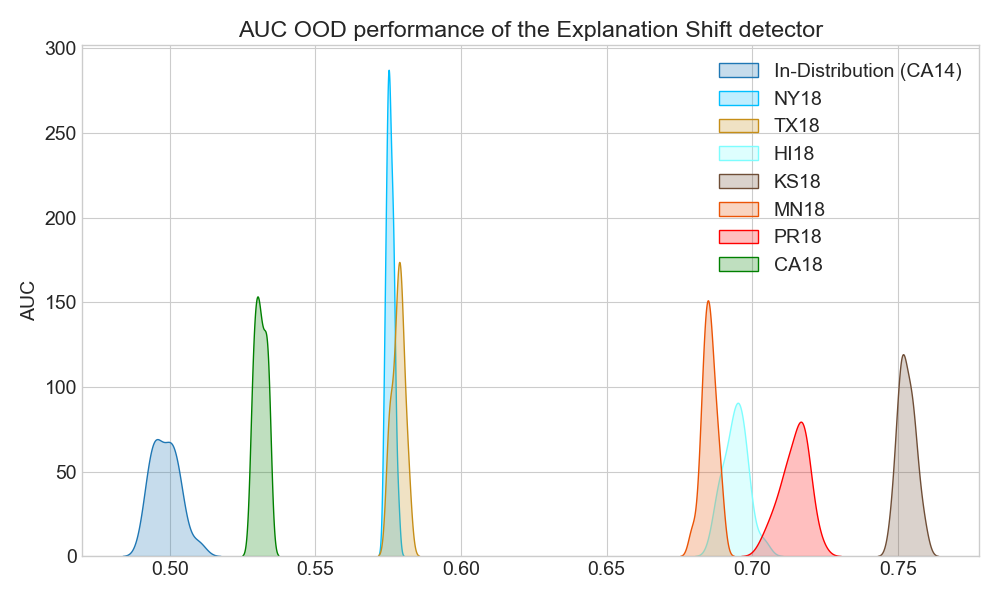}\hfill
\includegraphics[width=.49\textwidth]{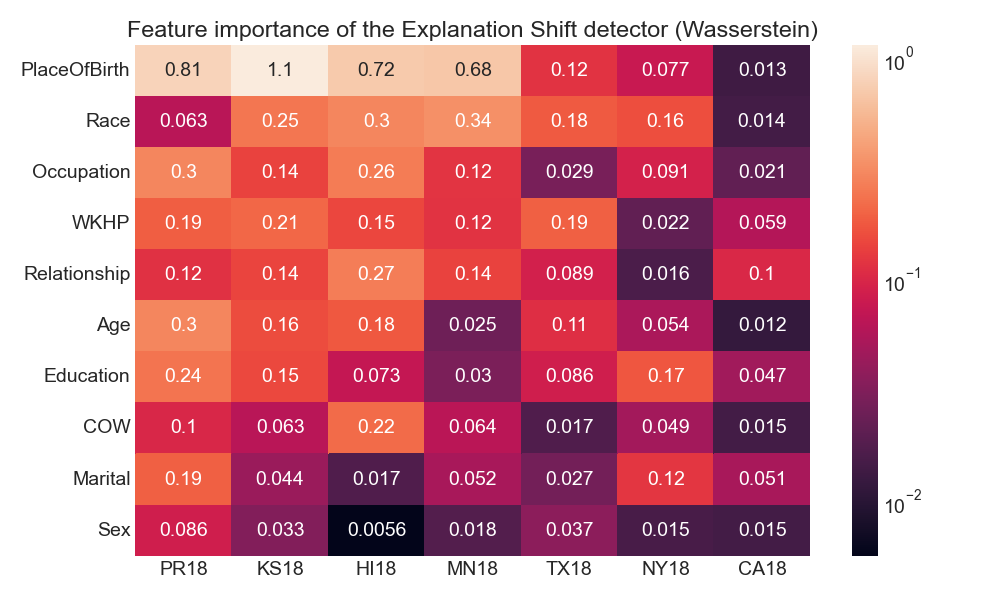}
\caption{In the left figure, comparison of the performance of \textit{Explanation Shift Detector}, in different states for the ACS Income prediction task.  In the left figure, we can see how the state with the highest OOD AUC detection is KS18 and not PR18 as in other prediction tasks, this difference with respect to the other prediction task can be attributed to \enquote{Place of Birth}, whose feature attributions the model finds to be more different than in CA14.}
\label{fig:xai.income}
\end{figure*}

\subsection{ACS Mobility}

The goal of this task is to predict whether an individual had the same residential address one year ago, only including individuals between the ages of 18 and 35. The goal of this filtering is to increase the prediction task difficulty, staying at the same address base rate is above $90\%$ for the population~\cite{ding2021retiring}.

The results of this experiment present a similar behaviour as the ACS Income prediction task (cf. Section \ref{fig:xai.income}), where the in-land states of the US are in an AUC range of $0.55-0.70$ and is the state of PR18 who achieves a higher OOD AUC. The features driving this behaviour are Citizenship for PR18 and Ancestry(Census record of your ancestors' lives with details like where they lived, who they lived with, and what they did for a living) for the other states. 

\begin{figure*}[ht]
\centering
\includegraphics[width=.49\textwidth]{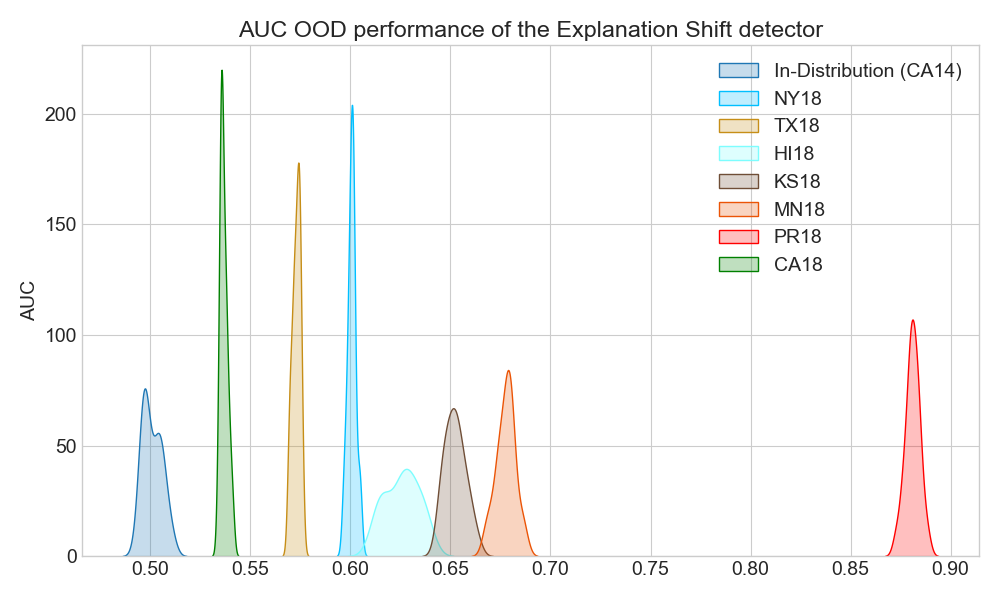}\hfill
\includegraphics[width=.49\textwidth]{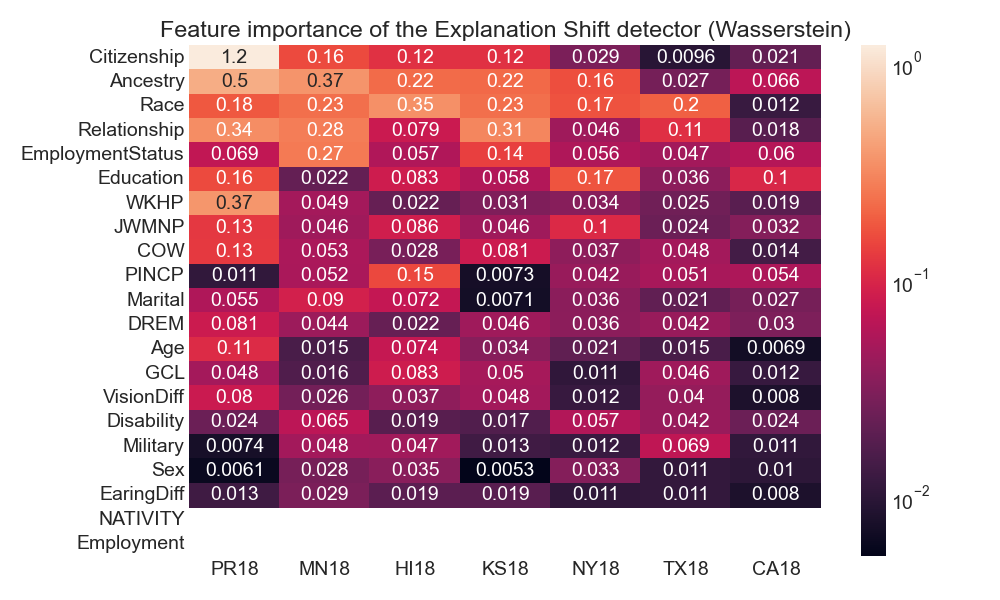}
\caption{In the left figure, comparison of the performance of \textit{Explanation Shift Detector}, in different states for the ACS Mobility dataset. Except for PR18, all the other states fall below an AUC of OOD detection of $0.70$. If we look at the features driving this difference is due to the Citizenship and the Ancestry relationship. For the other states protected social attributes such as Race or Marital status play an important role.}
\label{fig:xai.mobility}
\end{figure*}

%
%\section{You \emph{can} have an appendix here.}

%You can have as much text here as you want. The main body must be at most $8$ pages long.For the final version, one more page can be added.If you want, you can use an appendix like this one, even using the one-column format.
%%%%%%%%%%%%%%%%%%%%%%%%%%%%%%%%%%%%%%%%%%%%%%%%%%%%%%%%%%%%%%%%%%%%%%%%%%%%%%%
%%%%%%%%%%%%%%%%%%%%%%%%%%%%%%%%%%%%%%%%%%%%%%%%%%%%%%%%%%%%%%%%%%%%%%%%%%%%%%%

\end{document}